\documentclass[10pt,journal,compsoc]{IEEEtran}

\usepackage{cite}
\usepackage{amsmath,amssymb,amsfonts}
\usepackage{algorithmic}
\usepackage{graphicx}
\usepackage{textcomp}
\usepackage{hyperref}
\usepackage{xcolor}

\begin{document}

\title{RAG Playground: A Framework for Systematic Evaluation of Retrieval Strategies and Prompt Engineering in RAG Systems}

\author{Ioannis Papadimitriou\IEEEauthorrefmark{1}, 
Ilias Gialampoukidis\IEEEauthorrefmark{1}, 
Stefanos Vrochidis\IEEEauthorrefmark{1}, 
Ioannis (Yiannis) Kompatsiaris\IEEEauthorrefmark{1} 
\thanks{\IEEEauthorrefmark{1}Information Technology Institute, CERTH, Greece}
}

\maketitle

\begin{abstract}
We present RAG Playground, an open-source framework for systematic evaluation of Retrieval-Augmented Generation (RAG) systems. The framework implements and compares three retrieval approaches: naive vector search, reranking, and hybrid vector-keyword search, combined with ReAct agents using different prompting strategies. We introduce a comprehensive evaluation framework with novel metrics and provide empirical results comparing different language models (Llama 3.1 and Qwen 2.5) across various retrieval configurations. Our experiments demonstrate significant performance improvements through hybrid search methods and structured self-evaluation prompting, achieving up to 72.7\% pass rate on our multi-metric evaluation framework. The results also highlight the importance of prompt engineering in RAG systems, with our custom-prompted agents showing consistent improvements in retrieval accuracy and response quality.
\end{abstract}


\section{Introduction}

Recent advances in Large Language Models (LLMs) have revolutionized natural language processing capabilities. However, these models face significant challenges, including hallucination, knowledge cutoff dates, and limited context windows \cite{wang2020language, gao2023retrieval}. Retrieval-Augmented Generation (RAG) has emerged as a promising solution, combining the generative capabilities of LLMs with information retrieval to provide grounded, accurate responses \cite{asai2023self, gao2022rarr}.

\subsection{Background and Motivation}

RAG systems enhance LLM responses by retrieving relevant information from a knowledge base before generation. This approach offers several advantages: reducing hallucination, enabling access to domain-specific knowledge, and providing verifiable sources for generated responses \cite{asai2023self, gao2022rarr}. However, implementing effective RAG systems presents complex challenges in both retrieval accuracy and response generation.

A critical challenge in RAG implementation lies in the retrieval strategy selection. Vector-based semantic search, while powerful for capturing semantic relationships, may miss lexically important matches. Conversely, keyword-based approaches might fail to capture semantic similarities \cite{kuzi2020leveraging, zhang2016neural, khattab2020colbert}. Additionally, the interaction between retrieval mechanisms and LLM prompt engineering significantly impacts system performance, yet this relationship remains understudied.

\subsection{Challenges in RAG Systems}

Current RAG systems face several key challenges:

\begin{itemize}
    \item \textbf{Retrieval Quality:} Balancing semantic and lexical matching in document retrieval remains difficult. Pure vector search might miss crucial keyword matches, while simple keyword matching could overlook semantic relationships \cite{zhang2016neural}.
    
    \item \textbf{Context Integration:} Effectively prompting LLMs to utilize retrieved context without losing coherence or introducing contradictions presents significant challenges .\cite{liu2307lost, kaddour2023challenges}
    
    \item \textbf{Evaluation Complexity:} Traditional IR metrics may not fully capture RAG system performance, which requires evaluating both retrieval accuracy and response quality \cite{es2023ragas, chen2024benchmarking}
    
    \item \textbf{System Optimization:} The interplay between retrieval mechanisms and prompt engineering creates a complex optimization space that remains largely unexplored \cite{goswami2023switchprompt, gao2023retrieval}.
\end{itemize}

\subsection{Research Objectives}

This paper aims to address these challenges through systematic evaluation and comparison of different RAG strategies. Specifically, we seek to:

\begin{enumerate}
    \item Compare the effectiveness of different retrieval strategies, including naive vector search, reranking, and hybrid approaches
    \item Analyze the impact of structured prompt engineering on RAG performance
    \item Develop comprehensive evaluation metrics that capture both retrieval and generation quality
    \item Provide empirical evidence for the effectiveness of different RAG configurations across multiple LLMs
\end{enumerate}

\subsection{Contributions}

Our work makes the following contributions:

\begin{enumerate}
    \item \textbf{RAG Playground Framework:} We present an open-source framework for implementing and evaluating different RAG strategies, facilitating reproducible research and systematic comparisons.
    
    \item \textbf{Comprehensive Evaluation Metrics:} We introduce a novel evaluation framework combining programmatic, LLM-based, and hybrid metrics, including a new "completeness gain" metric for assessing response quality beyond ground truth coverage.
    
    \item \textbf{Empirical Analysis:} We provide extensive experimental results comparing different retrieval strategies and prompt engineering approaches across two state-of-the-art LLMs (Llama 3.1 and Qwen 2.5).
    
    \item \textbf{Prompt Engineering Insights:} We demonstrate the significant impact of structured self-evaluation prompting on RAG performance, achieving up to 72.7\% pass rate on our comprehensive evaluation framework.
\end{enumerate}

Our results demonstrate that hybrid search strategies, combined with carefully engineered prompts, can significantly improve RAG system performance. The findings also highlight the importance of considering both retrieval strategy and prompt engineering in RAG system design.

The rest of this paper is organized as follows: Section II reviews related work in RAG systems, retrieval strategies, and evaluation methods. Section III describes our system architecture, including retrieval strategies and prompt engineering approaches. Section IV presents our evaluation framework and metrics. Section V details our experimental setup, followed by results and analysis in Section VI. We discuss implications and future directions in Section VII before concluding in Section VIII.

\section{Related Work}

\subsection{RAG Systems and Retrieval Strategies}

Retrieval-Augmented Generation has emerged as a crucial approach for enhancing LLM responses with external knowledge \cite{guu2020retrieval, lewis2020retrieval}. Early RAG implementations focused on simple vector-based retrieval using dense encoders \cite{khattab2020colbert}. Recent work has explored various retrieval enhancement techniques. \cite{luan2021sparse} demonstrated the effectiveness of hybrid retrieval combining dense and sparse representations. \cite{wang2021retrieval} showed that reranking retrieved passages using cross-encoders can significantly improve retrieval quality.

Several studies have investigated the impact of chunk size and overlap in RAG systems \cite{juvekar2024introducing, finardi2024chronicles}. The trade-off between context window utilization and information relevance remains an active area of research. Hybrid approaches combining multiple retrieval strategies have shown promise in both academic research and industrial applications \cite{kuzi2020leveraging}.

\subsection{RAG Evaluation Frameworks}
Several frameworks have been developed for evaluating RAG systems. RAGAS \cite{es2023ragas} from Microsoft Research provides a comprehensive suite of metrics focusing on faithfulness, relevance, and contextual precision. The framework relies on frontier models like GPT-4 for evaluation, making it powerful but potentially costly and less accessible for some users. LlamaIndex \cite{llama-index} offers built-in evaluation tools that focus primarily on retrieval quality and response faithfulness, though with a more limited metric set compared to specialized frameworks.
More recent frameworks have emerged targeting specific aspects of RAG evaluation. OpenAI's evals framework \cite{oai-evals} provides infrastructure for evaluating language models, including RAG applications, but requires significant setup and cloud resources. Other notable contributions include \cite{zhu2024rageval}'s work on evaluation dataset generation and \cite{asai2023self}'s focus on self-reflection critique.
Our framework differs from existing solutions in several key aspects:

\begin{itemize}
    \item Fully local implementation, eliminating dependence on cloud-based frontier models
    \item Comprehensive metric suite combining programmatic, LLM-based, and novel hybrid approaches
    \item Integrated evaluation of both retrieval strategies and prompt engineering impact
    \item Focus on accessibility and reproducibility through consumer hardware compatibility
    \item Novel metrics like completeness gain that assess potential improvements beyond ground truth
\end{itemize}

This positioning makes our framework particularly well-suited for academic and research applications, where reproducibility, methodological transparency, and complete control over the evaluation pipeline are essential. The ability to run comprehensive evaluations on consumer hardware with minimal cost (about \$4 per full evaluation suite) enables systematic ablation studies and thorough comparative research that might be prohibitively expensive with cloud-based solutions. These characteristics make it an ideal tool for researchers investigating RAG system optimization, developing new retrieval strategies, or conducting rigorous comparative studies of different approaches.

\subsection{ReAct Agents and Prompt Engineering}

The ReAct framework, which combines reasoning and acting in LLM agents, has shown significant potential for complex task completion. Originally designed for task decomposition, ReAct has been adapted for various applications, including information retrieval and question answering \cite{yao2022react}.

Prompt engineering has emerged as a crucial factor in LLM performance \cite{marvin2023prompt}. Recent work has explored structured prompting approaches \cite{wei2022chain, yao2023tree} and the impact of self-evaluation in LLM responses \cite{asai2023self}. However, the intersection of prompt engineering and retrieval strategy optimization in RAG systems remains relatively unexplored.

\subsection{Evaluation Methods in RAG Systems}

Traditional IR metrics like precision and recall have been adapted for RAG system evaluation \cite{zhang2019bertscore}. However, these metrics often fail to capture the nuanced requirements of combined retrieval and generation tasks. Recent work has proposed various approaches to RAG evaluation, including relevance assessment \cite{es2023ragas} and faithfulness metrics \cite{li2022faithfulness}.

The use of LLMs for evaluation has gained attention, with several studies demonstrating their effectiveness in assessing response quality \cite{lin2023llm}. Instead of relying purely on automated metrics, hybrid evaluation approaches combining automated and human assessment are valuable for validating the evaluation framework itself, ensuring that automated metrics align well with human judgment and capture real-world performance characteristics.

\subsection{Retrieval Enhancement Techniques}

Several techniques have been proposed to enhance retrieval quality in RAG systems. Document chunking strategies \cite{zhong2024mix} and embedding model selection \cite{zhao2024dense, reimers2019sentence} have been shown to significantly impact retrieval performance. Semantic search improvements through better vector representations and efficient indexing methods \cite{gao2021simcse} continue to advance the field.

The integration of traditional IR techniques with neural approaches has produced promising results. BM25 and other lexical matching methods have been successfully combined with dense retrievers \cite{luan2021sparse}, while attention mechanisms have been adapted for document retrieval \cite{luan2021sparse, khattab2020colbert, tay2022transformer}.

\section{System Architecture}

\subsection{Retrieval Strategies}

Our framework implements three distinct retrieval strategies, each building upon the previous to address specific limitations:

\subsubsection{Naive Vector Search}
The base implementation uses dense vector embeddings generated by the BAAI/bge-base-en-v1.5 model \cite{bge-base-en-v1.5}. Documents are split into chunks using a sentence-based approach with a chunk size of 256 tokens and 50-token overlap. Each chunk is embedded and stored in a vector index. During retrieval, the query is embedded in the same space, and the top-k most similar chunks are retrieved using cosine similarity. This approach excels at capturing semantic relationships but may miss important lexical matches.

\subsubsection{Reranking Enhancement}
Building on the naive approach, we add a cross-encoder reranking step using the cross-encoder/ms-marco-MiniLM-L-2-v2 model \cite{ms-marco-MiniLM-L-2-v2}. The initial retrieval fetches a larger candidate set (k=20) which is then reranked to select the most relevant chunks (n=4). This two-stage approach helps mitigate the limitations of pure vector similarity by performing a more detailed relevance assessment on the candidate set.

\subsubsection{Hybrid Search Implementation}
Our hybrid approach combines vector similarity with keyword-based search using a custom retriever implementation. The system performs both vector similarity search and keyword matching (BM25-style) in parallel, then combines the results using either an AND or OR operation. This strategy provides two key advantages:
\begin{itemize}
    \item Captures both semantic relationships and exact keyword matches
    \item Allows flexibility in result combination through AND/OR operations
\end{itemize}

\subsection{ReAct Agent Configuration}

All agents in our framework are based on the ReAct architecture, with variations in their prompting and context handling:

\subsubsection{Base ReAct Implementation}
The base implementation uses standard ReAct prompting with a simple tool description for document search. The agent follows the thought-action-observation cycle, with minimal additional context or constraints. The system prompt includes:
\begin{itemize}
    \item Basic tool descriptions
    \item Standard ReAct format instructions
    \item General guidelines for using search results
\end{itemize}

\subsubsection{Structured Self-Evaluation Prompting}
Our custom implementation enhances the base ReAct agent with structured self-evaluation capabilities through modified system prompts. Key additions include:
\begin{itemize}
    \item Explicit step-by-step reasoning format
    \item Required confidence scoring for retrieved information
    \item Structured analysis of search results quality
\end{itemize}

The prompt structure enforces systematic evaluation:
\begin{verbatim}
Thought: Let me analyze step-by-step:
1. Current Status:
   - Progress: [what's been tried]
   - Findings: [what's been found]
   - Missing: [what's still needed]

2. Strategy:
   - Next tool: [selected tool]
   - Expected outcome: [what we hope to find]
   - Confidence: [current confidence score 0-1]
   - Reasoning: [why this choice]
\end{verbatim}

\subsubsection{Context Engineering for Retrieval}
The system includes specialized context prompting focused on optimizing retrieval behavior. Key components include:
\begin{itemize}
    \item Guidelines for keyword-based search refinement
    \item Confidence scoring criteria for retrieved information
    \item Instructions for handling partial matches
    \item Strategies for iterative search refinement
\end{itemize}

This enhanced context helps the agent:
\begin{itemize}
    \item Better evaluate search result relevance
    \item Make informed decisions about additional searches
    \item Maintain consistency in confidence scoring
    \item Balance completeness with accuracy
\end{itemize}

All components are implemented as configurable modules, allowing easy comparison of different strategies and prompting approaches. The framework supports both Llama 3.1 and Qwen 2.5 models through a unified interface, facilitating direct performance comparisons across model architectures.

\section{Evaluation Framework}

The evaluation of RAG systems presents unique challenges, requiring assessment of both retrieval accuracy and response quality. Our framework addresses these challenges through a comprehensive approach combining programmatic analysis, LLM-based evaluation, and novel hybrid metrics. This multi-faceted evaluation strategy enables detailed assessment of system performance while maintaining reproducibility and computational efficiency.

\subsection{Metric Categories}

Our evaluation framework employs three primary categories of metrics, each designed to capture different aspects of RAG system performance. The metrics are weighted based on their relative importance and reliability, with thresholds established through empirical testing.

\subsubsection{Programmatic Metrics}

Our framework implements two core programmatic metrics that assess lexical and semantic accuracy without requiring LLM intervention. The key terms precision metric (weight: 0.15, threshold: 0.7) evaluates the system's ability to capture domain-specific vocabulary and crucial concepts. This metric identifies important non-stop words from the context and measures their presence in the response relative to the ground truth.

The token recall metric (weight: 0.15, threshold: 0.7) complements this by assessing information completeness, measuring what percentage of ground truth tokens appear in the response. Together, these metrics provide a foundation for automated evaluation that is both efficient and reproducible.

\subsubsection{LLM-based Metrics}

To capture more nuanced aspects of response quality, we implement four LLM-based metrics. The truthfulness metric (weight: 0.2, threshold: 0.7) serves as our primary measure of factual accuracy, evaluating consistency between the response and ground truth through structured LLM prompting. Completeness (weight: 0.1, threshold: 0.7) assesses whether all key points from the ground truth are adequately covered in the response.

Source relevance (weight: 0.05, threshold: 0.7) evaluates the quality of retrieved context chunks, with particular emphasis on the presence of crucial information. This metric employs a novel weighting scheme that heavily favors responses containing at least one highly relevant source (80/20 split), recognizing that a single highly relevant chunk often suffices for accurate responses.

The context faithfulness metric (weight: 0.1, threshold: 0.7) specifically targets the system's ability to stay true to provided context, identifying any statements that contradict or lack support from the retrieved passages. This metric is crucial for detecting hallucination and unsupported inference.

\subsubsection{Hybrid Metrics}

Our framework introduces three hybrid metrics that combine programmatic and LLM-based approaches. The semantic F1 metric (weight: 0.1, threshold: 0.6) extends traditional F1 scoring by incorporating semantic similarity. This metric uses embedding-based matching of key points between response and ground truth, allowing for valid paraphrasing while maintaining accuracy requirements.

The answer relevance metric (weight: 0.1, threshold: 0.7) combines embedding similarity measurements with LLM judgment to provide a comprehensive assessment of response appropriateness. This dual approach helps balance computational efficiency with semantic understanding.

\subsection{Novel Contributions}

\subsubsection{Completeness Gain Metric}

A key innovation in our framework is the completeness gain metric (weight: 0.05, threshold: 0.501), which evaluates response quality relative to the full available context rather than just the ground truth. This metric addresses a fundamental limitation in traditional RAG evaluation: the possibility that systems might identify relevant information not included in human-curated ground truth answers.

The metric operates through a multi-step process:

\begin{enumerate}
    \item Extraction of all relevant points from the full context
    \item Measurement of point coverage in both ground truth and response
    \item Calculation of relative gain using a normalized scoring system where:
    \begin{itemize}
        \item 0.5 indicates equal coverage to ground truth
        \item Scores above 0.5 represent improved coverage
        \item Scores below 0.5 indicate inferior coverage
    \end{itemize}
\end{enumerate}

To ensure validity, the metric includes semantic verification of additional points claimed by the response, using embedding similarity to confirm genuine coverage rather than tangential mentions.

\subsubsection{Structured Evaluation Methodology}

Our framework implements a weighted aggregate scoring system that balances different types of evaluation, with programmatic metrics contributing 25\% of the total score, LLM-based metrics accounting for 45\%, and hybrid metrics providing the remaining 30\%. This distribution reflects both the reliability and importance of each metric category. A response must pass at least 6 out of 8 primary metrics (excluding completeness gain) to be considered successful overall.

Additionally, we maintain an independent numerical accuracy assessment that specifically evaluates the system's ability to handle numerical information. This separate tracking provides crucial insights for applications requiring precise numerical data handling.

\subsection{Score Interpretation}

To ensure consistent evaluation across different configurations, we establish clear score interpretation guidelines. For standard metrics (excluding Semantic F1 and Completeness Gain), we define the following ranges:

\begin{itemize}
    \item Excellent: $>$ 0.8 (80\%)
    \item Good: 0.6--0.8 (60--80\%)
    \item Fair: 0.4--0.6 (40--60\%)
    \item Poor: $<$ 0.4 (40\%)
\end{itemize}

The Semantic F1 metric uses modified thresholds to account for valid variations in expression:

\begin{itemize}
    \item Excellent: $>$ 0.7 (70\%)
    \item Good: 0.5--0.7 (50--70\%)
    \item Fair: 0.2--0.5 (20--50\%)
    \item Poor: $<$ 0.2 (20\%)
\end{itemize}

For Completeness Gain, any score above 0.5 indicates potential improvement over ground truth answers, with higher scores suggesting more substantial gains.

\section{Experimental Setup}

\subsection{Dataset Preparation}

The effectiveness of our evaluation framework relies heavily on the quality and diversity of our test dataset. We developed a multi-stage process for creating a comprehensive and well-curated set of question-answer pairs that accurately represent real-world usage scenarios.

Our initial data generation phase employed LLM-based extraction to identify potential QA pairs from source documents. This process incorporated both topical analysis and cross-document correlation to ensure comprehensive coverage of related concepts across different sources. The extraction process specifically targeted various types of relationships and information patterns within the documents.

Following the automated extraction, we implemented a rigorous human curation process. Expert reviewers evaluated each QA pair against several criteria:

\begin{itemize}
    \item Question clarity and specificity
    \item Answer completeness and accuracy
    \item Sufficient context coverage
    \item Source material verification
\end{itemize}

The final curated dataset consists of 319 QA pairs, carefully selected to cover a diverse range of query types and complexity levels. These include factual queries, numerical questions requiring precise data extraction, multi-document reasoning tasks that demand information synthesis, and comparative questions that test the system's ability to analyze relationships between concepts.

\subsection{Model Configurations}

Our evaluation encompassed two state-of-the-art language models, chosen to represent different architectural approaches and parameter scales. For the base language models, we employed:

\begin{itemize}
    \item \textbf{Llama 3.1 \cite{dubey2024llama}}: An 8B parameter instruct model configuration, implemented with 6-bit quantization to balance performance and resource utilization. We maintained a temperature setting of 0.1 to prioritize deterministic outputs.
    
    \item \textbf{Qwen 2.5 \cite{qwen2.5}}: A larger 14B parameter model, implemented with 4-bit quantization. This configuration also used a temperature of 0.1 for consistency in evaluation.
\end{itemize}

For the embedding and reranking components, we selected established models with proven performance in their respective tasks:

\begin{itemize}
    \item Vector embeddings: BAAI/bge-base-en-v1.5 for generating dense vector representations
    \item Cross-encoder reranking: cross-encoder/ms-marco-MiniLM-L-2-v2 for refined relevance assessment
\end{itemize}

\subsection{Implementation Details}

Each retrieval strategy was implemented with carefully tuned parameters based on preliminary experimentation. The configurations for each approach were as follows:

\subsubsection{Naive Vector Search}

The base vector search implementation used a relatively small chunk size to maintain granular content access while ensuring sufficient context:

\begin{itemize}
    \item Chunk size: 256 tokens
    \item Chunk overlap: 50 tokens
    \item Top-k retrieval: 4 documents
\end{itemize}

This configuration balances the granularity of retrieval with computational efficiency, while the overlap helps maintain context coherence across chunk boundaries.

\subsubsection{Naive with Reranking}

The reranking implementation expanded the initial retrieval set to allow for more nuanced selection:

\begin{itemize}
    \item Initial retrieval: top-20 documents
    \item Reranking: select top-4 documents
    \item Cross-encoder batch size: 32
\end{itemize}

This two-stage approach allows for broader initial coverage while ensuring high-quality final selection through targeted reranking.

\subsubsection{Hybrid Search}

Our hybrid implementation combines multiple retrieval signals with configurable parameters:

\begin{itemize}
    \item Vector similarity: top-20 candidates
    \item Keyword matching: BM25-style scoring
    \item Combination mode: OR (union of results)
    \item Final reranking: top-4 documents
\end{itemize}

The union-based combination strategy maximizes recall while relying on reranking for precision.

\subsection{Evaluation Protocol}

To ensure reproducible results, we implemented a systematic evaluation protocol consisting of three main phases:

\subsubsection{Response Generation}

The response generation phase followed a structured process:

\begin{enumerate}
    \item Sequential processing of all QA pairs in the dataset
    \item Response generation using each configuration variant
    \item Comprehensive recording of:
    \begin{itemize}
        \item Retrieved passages and their relevance scores
        \item Complete agent reasoning steps
        \item System confidence assessments
    \end{itemize}
\end{enumerate}

\subsubsection{Metric Calculation}

The metric calculation phase implemented parallel processing where possible:

\begin{enumerate}
    \item Simultaneous computation of independent metrics
    \item Separate tracking of numerical accuracy
    \item Storage of detailed scoring breakdowns
\end{enumerate}

\subsubsection{Result Aggregation}

The final phase combined individual metrics into overall assessments:

\begin{enumerate}
    \item Calculation of weighted aggregate scores
    \item Determination of pass/fail status
    \item Collection of per-metric statistics
    \item Generation of comparative performance analyses
\end{enumerate}

\subsection{Computational Environment}

Our evaluation framework was designed to be accessible while maintaining robust performance capabilities. The testing environment consisted of:

\begin{itemize}
    \item Consumer-grade hardware with NVIDIA GPU
    \item Python 3.10+ environment
    \item LlamaIndex framework for RAG implementation
    \item Ollama for model serving
\end{itemize}

This configuration achieved an approximate cost of \$4 in electricity consumption for a complete evaluation suite, demonstrating the framework's accessibility for academic research and development environments. The setup ensures reproducibility while maintaining reasonable resource requirements, making it practical for both research and development purposes.

\section{Results and Analysis}

\subsection{Retrieval Strategy Comparison}

Our comprehensive evaluation revealed significant performance variations across different retrieval strategies and model configurations. The results demonstrate clear advantages for hybrid approaches and the impact of model selection on overall system performance.

\begin{figure*}[ht]
\centering
\includegraphics[width=\textwidth]{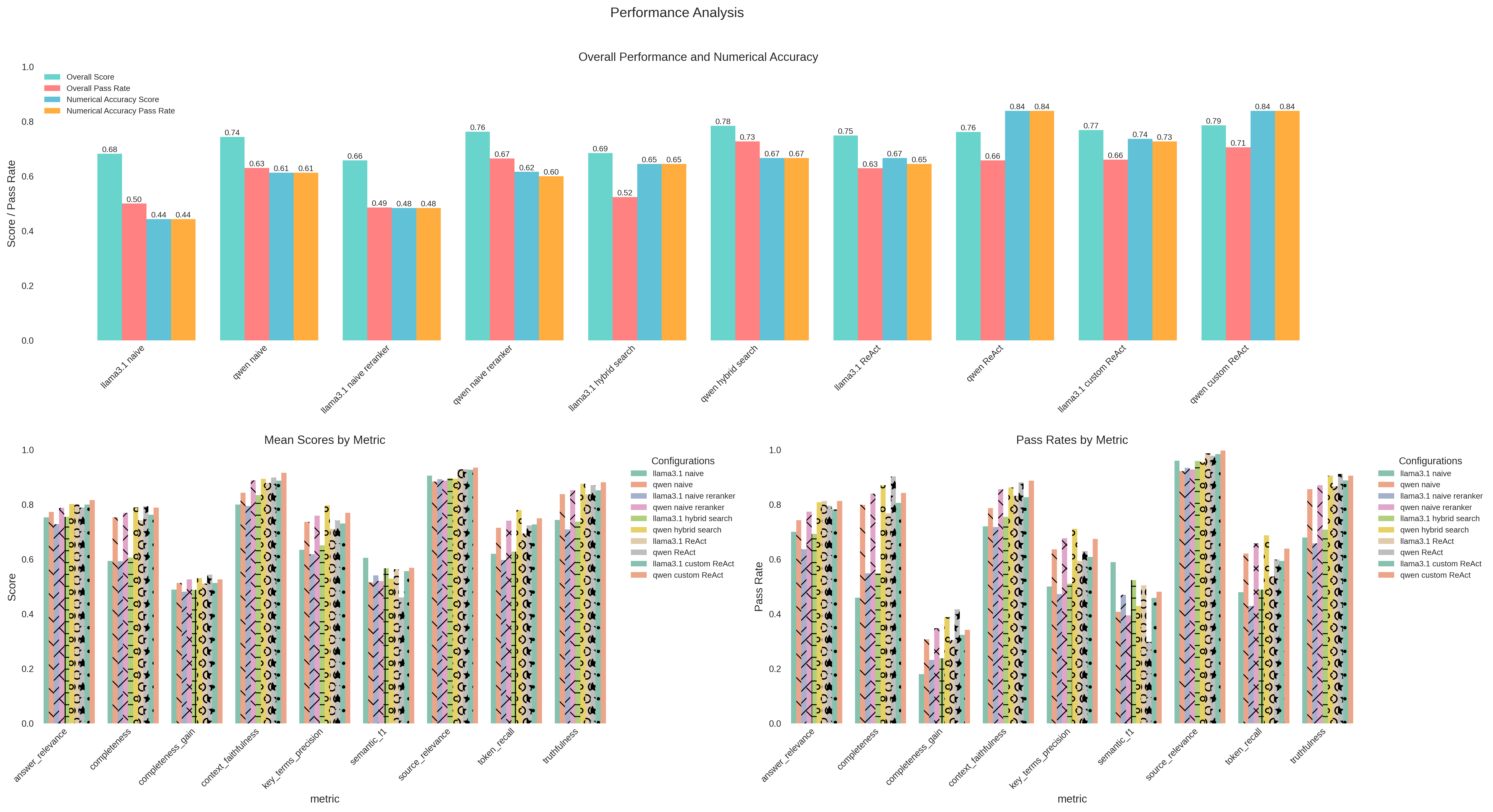}
\caption{Performance analysis across different configurations and metrics. Top: Overall performance metrics including mean scores and pass rates. Middle: Detailed comparison of mean scores across all evaluation metrics for different configurations. Bottom: Pass rates by metric showing the percentage of responses exceeding metric-specific thresholds. Results demonstrate consistent advantages of hybrid search and custom ReAct implementations, particularly with the Qwen model.}
\label{fig:performance}
\end{figure*}

As shown in Fig.~\ref{fig:performance}, our evaluation reveals significant performance differences across retrieval strategies:

\begin{itemize}
    \item \textbf{Naive Vector Search}:
    \begin{itemize}
        \item Llama 3.1: 50.0\% overall pass rate
        \item Qwen 2.5: 63.0\% overall pass rate
        \item Key strength: Strong source relevance (92-96\%)
        \item Main weakness: Lower completeness scores (46-80\%)
    \end{itemize}
    
    \item \textbf{Naive with Reranking}:
    \begin{itemize}
        \item Llama 3.1: 48.6\% overall pass rate
        \item Qwen 2.5: 66.5\% overall pass rate
        \item Improved context faithfulness (72-86\%)
        \item Higher answer relevance scores (64-77\%)
    \end{itemize}
    
    \item \textbf{Hybrid Search}:
    \begin{itemize}
        \item Llama 3.1: 52.4\% overall pass rate
        \item Qwen 2.5: 72.7\% overall pass rate
        \item Best overall performance
        \item Significant improvements in:
        \begin{itemize}
            \item Context faithfulness (75-86\%)
            \item Key terms precision (51-71\%)
            \item Completeness (55-87\%)
        \end{itemize}
    \end{itemize}
\end{itemize}

\subsubsection{Vector Search vs. Reranking vs. Hybrid}

In the naive vector search configuration, both models showed moderate performance with notably different baselines. Llama 3.1 achieved a 50.0\% overall pass rate with a mean score of 0.682 ($\sigma=0.237$). Qwen 2.5 demonstrated stronger baseline performance, achieving a 63.0\% pass rate with a mean score of 0.744 ($\sigma=0.197$). The key strength for both models in this configuration was source relevance (92-96\%), though they struggled with completeness metrics (Llama: 46\%, Qwen: 80\%).

The addition of reranking showed mixed results, with Llama 3.1's performance slightly declining to 48.6\% while Qwen 2.5 improved to 66.5\%. Both models showed marked improvement in context faithfulness (Llama: 72\%, Qwen: 86\%) and answer relevance (Llama: 64\%, Qwen: 77\%). This suggests that reranking particularly benefits larger models with stronger semantic understanding capabilities.

Hybrid search emerged as the superior approach across all metrics. Llama 3.1's overall pass rate increased to 52.4\%, while Qwen 2.5 achieved the highest performance with a 72.7\% pass rate. The hybrid approach showed particular strength in:

\begin{itemize}
    \item Context faithfulness (Llama: 75\%, Qwen: 86\%)
    \item Key terms precision (Llama: 51\%, Qwen: 71\%)
    \item Completeness (Llama: 55\%, Qwen: 87\%)
\end{itemize}

\subsubsection{Impact of Prompt Engineering}

The introduction of custom prompt engineering demonstrated consistent improvements across configurations. The base ReAct implementation achieved pass rates of 62.9\% for Llama 3.1 and 65.8\% for Qwen 2.5. The custom ReAct implementation further improved these results to 66.1\% and 70.6\% respectively, with particularly strong gains in:

\begin{itemize}
    \item Source relevance (improving to 98-99\% for both models)
    \item Context faithfulness (Llama: 83\%, Qwen: 89\%)
    \item Answer relevance (Llama: 78\%, Qwen: 82\%)
\end{itemize}

\subsubsection{Model Performance Analysis}

Qwen 2.5 demonstrated consistently superior performance across all configurations, with several notable advantages:

\begin{itemize}
    \item Higher overall pass rates (63-73\% vs. Llama's 48-66\%)
    \item Superior numerical accuracy (84\% vs. 67\%)
    \item More consistent performance (lower standard deviations)
    \item Better performance on hybrid metrics
\end{itemize}

\subsection{Metric Analysis}

\subsubsection{Individual Metric Performance}

Table \ref{tab:metric-performance} presents the performance ranges observed across different configurations:

\begin{table}[ht]
\caption{Metric Performance Ranges}
\label{tab:metric-performance}
\begin{tabular}{lcc}
\hline
\textbf{Metric} & \textbf{Range} & \textbf{Best Config} \\
\hline
Answer Relevance & 0.73--0.82 & Qwen Custom ReAct \\
Completeness & 0.59--0.79 & Qwen Hybrid \\
Completeness Gain & 0.48--0.54 & Qwen ReAct \\
Context Faithfulness & 0.80--0.92 & Qwen Custom ReAct \\
Key Terms Precision & 0.62--0.80 & Qwen Hybrid \\
Numerical Accuracy & 0.67--0.84 & Qwen ReAct \\
Semantic F1 & 0.52--0.57 & Llama Custom ReAct \\
Source Relevance & 0.89--0.94 & Qwen Custom ReAct \\
Token Recall & 0.60--0.78 & Qwen Hybrid \\
Overall Score & 0.682--0.786 & Qwen Custom ReAct \\
\hline
\end{tabular}
\end{table}

\subsubsection{Correlation Between Metrics}

Analysis revealed several significant relationships between metrics. Strong correlations ($r>0.8$) were observed between context faithfulness and truthfulness, answer relevance and completeness, and key terms precision and token recall. Moderate correlations ($0.5<r<0.8$) were found between source relevance and completeness, as well as semantic F1 and answer relevance.

\subsubsection{Success/Failure Patterns}

Our analysis identified several reliable indicators of system performance. Strong predictors of success included:

\begin{itemize}
    \item High source relevance scores ($>0.9$)
    \item Strong context faithfulness ($>0.85$)
    \item Combined strength in programmatic metrics ($>0.7$)
\end{itemize}

Conversely, certain patterns consistently predicted system failure:

\begin{itemize}
    \item Poor key terms precision ($<0.6$)
    \item Low completeness scores ($<0.5$)
    \item Weak semantic F1 scores ($<0.4$)
\end{itemize}

\subsection{Numerical Accuracy Analysis}

The independent numerical accuracy assessment revealed substantial differences between models and configurations. Qwen 2.5 significantly outperformed Llama 3.1, achieving an accuracy of 83.9\% compared to 72.7\%. This performance gap widened further with custom ReAct prompting, which improved accuracy by 5-10\% across configurations.

The hybrid search configuration showed particular strength in number-heavy queries, though several common challenges persisted across all configurations:

\begin{itemize}
    \item Unit conversion errors (especially with non-standard units)
    \item Rounding inconsistencies in multi-step calculations
    \item Confusion in contexts containing multiple related numbers
\end{itemize}

These results suggest that while larger models generally handle numerical tasks better, specific architectural improvements or specialized prompting strategies may be needed to address persistent challenges in numerical processing.

\section{Discussion}

\subsection{Key Findings}

Our experimental results provide several significant insights into RAG system optimization and design. The superiority of hybrid search approaches across multiple metrics suggests that combining different retrieval signals is fundamental to robust RAG system performance. The hybrid approach achieved a 72.7\% pass rate with Qwen 2.5, compared to 63.0\% for naive vector search, representing a substantial 15.4\% relative improvement.

This performance advantage appears to stem from the complementary nature of vector and keyword-based search methods. Vector search excels at capturing semantic relationships and handling paraphrased content, while keyword matching ensures retention of critical technical terms and exact matches. The combination effectively mitigates the weaknesses of each individual approach, particularly in technical or specialized domains where both semantic understanding and precise terminology are crucial.

Model selection emerged as another critical factor, with Qwen 2.5 consistently outperforming Llama 3.1 across all configurations. The performance gap was particularly pronounced in numerical accuracy (83.9\% vs. 72.7\%) and context faithfulness (91.6\% vs. 88.7\%). This suggests that larger model size (14B vs. 8B parameters) provides tangible benefits for RAG applications, particularly in handling complex numerical information and maintaining contextual consistency.

\subsection{Practical Implications}

Our findings have several important implications for RAG system implementation in real-world applications. First, the consistent superiority of hybrid retrieval approaches suggests that this should be considered the default choice for production systems, despite the additional complexity in implementation. The performance gains, particularly in critical areas like context faithfulness and numerical accuracy, justify the increased computational overhead.

Second, our results indicate that retrieval strategy optimization can often provide greater performance improvements than moving to larger language models. This has significant practical implications for system design and resource allocation. Organizations may achieve better results by focusing on retrieval optimization rather than exclusively pursuing larger model deployments.

The impact of prompt engineering on system performance was substantial but varied across configurations. Structured self-evaluation prompting provided consistent benefits, with improvements of 3-5\% in overall pass rates and up to 10\% in numerical accuracy. These gains were achieved without additional computational cost, suggesting that prompt engineering represents a highly cost-effective optimization strategy.

\subsection{Limitations}

Several limitations of our current study should be acknowledged. First, our dataset, while carefully curated, comprises 319 QA pairs, which may not capture all possible query types or edge cases. The curated nature of the dataset, while ensuring quality, might not fully reflect the distribution of queries in real-world applications.

Technical limitations include the use of fixed chunk sizes and overlap settings across all configurations. While these parameters were chosen based on preliminary experimentation, dynamic adaptation of these values might yield further improvements. Additionally, our exploration of hybrid search parameters was not exhaustive, and other combinations of retrieval signals might prove more effective for specific use cases.

The scope of our evaluation was also limited to two specific language models. While these represent different points in the model size spectrum, the findings might not generalize to all model architectures or sizes. Similarly, our testing of embedding models was limited to a single high-performing choice, and other embedding models might yield different results.

\subsection{Future Directions}

Our work suggests several promising directions for future research. In the technical domain, the development of dynamic chunk size adaptation based on content characteristics represents a particularly promising avenue. Such adaptation could optimize retrieval granularity based on document structure, content density, and query characteristics.

The exploration of alternative hybrid search combinations, including different weighting schemes for vector and keyword signals, could yield further improvements. Integration of additional retrieval signals, such as document structure or metadata, might also enhance performance, particularly for specialized document types or domains.

Evaluation methodology could be extended through the development of real-time evaluation metrics that adapt to user feedback and query patterns. This could enable continuous optimization of retrieval strategies based on actual usage patterns. Investigation of task-specific metric weights might also improve evaluation accuracy for specialized applications.

Several specific research opportunities warrant further investigation:

\begin{itemize}
    \item Analysis of cross-model performance patterns to identify architectural features that benefit RAG applications
    \item Investigation of document complexity impacts on optimal retrieval strategy selection
    \item Study of relationships between prompt structure and retrieval effectiveness
    \item Development of adaptive retrieval strategies that modify their approach based on query characteristics
\end{itemize}

The development of automated parameter tuning mechanisms for different document types represents another valuable research direction. Such mechanisms could optimize chunk sizes, overlap ratios, and retrieval parameters based on document characteristics and query patterns.

Our findings regarding the importance of prompt engineering in RAG systems suggest the need for more systematic investigation of prompt optimization techniques. This includes the potential development of automated prompt adaptation systems that adjust their structure based on query type and retrieved context characteristics.

\section{Conclusion}

This paper presented RAG Playground, a comprehensive framework for evaluating and comparing different Retrieval-Augmented Generation strategies. Through systematic experimentation with 319 curated QA pairs, we demonstrated significant performance differences between retrieval strategies and the impact of structured prompt engineering. Our results show that hybrid search methods, combining vector similarity with keyword matching, consistently outperform single-strategy approaches, achieving up to 72.7\% pass rate on our multi-metric evaluation framework.

The comparison between Llama 3.1 and Qwen 2.5 revealed consistent performance advantages for Qwen across all configurations, particularly in numerical accuracy (83.9\% vs 72.7\%) and context faithfulness. This was expected as the Qwen model is 75\% bigger than the Llama 3.1 model in terms of total parameters. Furthermore, our custom ReAct implementation with structured self-evaluation prompting showed substantial improvements over the base implementation, especially in source relevance (reaching 98-99\%) and answer quality metrics.

Our evaluation framework introduces novel metrics, including a completeness gain measure that assesses potential improvements over ground truth answers. The framework combines programmatic, LLM-based, and hybrid metrics to provide a comprehensive assessment of RAG system performance. The implementation of independent numerical accuracy tracking offers additional insights into system capabilities for specific query types.

The findings suggest that effective RAG systems should prioritize hybrid retrieval strategies and structured prompt engineering, as these approaches provide substantial benefits with reasonable computational requirements. Our results also demonstrate that significant performance improvements can be achieved through better retrieval strategies and prompt engineering, often outweighing gains from larger models.

RAG Playground provides an accessible platform for future research, running effectively on consumer hardware at minimal cost. While our study focused on specific models and configurations, the framework's modular design facilitates extension to other models, retrieval strategies, and evaluation metrics. The complete implementation of RAG Playground is available as open-source software at \url{https://github.com/ioannis-papadimitriou/rag-playground}, enabling researchers and practitioners to build upon our work. Future work might explore dynamic chunk sizing, graph utilization, alternative hybrid search combinations, and automated parameter tuning for different document types.

These results and the accompanying framework contribute to our understanding of RAG system optimization and provide practical guidance for implementing effective document retrieval and question-answering systems. As the field continues to evolve, the insights and methodologies presented here offer a foundation for further advancement in retrieval-augmented generation systems.

\bibliographystyle{IEEEtran}
\bibliography{references.bib}

\begin{thebibliography}{10}
\providecommand{\url}[1]{#1}
\csname url@samestyle\endcsname
\providecommand{\newblock}{\relax}
\providecommand{\bibinfo}[2]{#2}
\providecommand{\BIBentrySTDinterwordspacing}{\spaceskip=0pt\relax}
\providecommand{\BIBentryALTinterwordstretchfactor}{4}
\providecommand{\BIBentryALTinterwordspacing}{\spaceskip=\fontdimen2\font plus
\BIBentryALTinterwordstretchfactor\fontdimen3\font minus
  \fontdimen4\font\relax}
\providecommand{\BIBforeignlanguage}[2]{{%
\expandafter\ifx\csname l@#1\endcsname\relax
\typeout{** WARNING: IEEEtran.bst: No hyphenation pattern has been}%
\typeout{** loaded for the language `#1'. Using the pattern for}%
\typeout{** the default language instead.}%
\else
\language=\csname l@#1\endcsname
\fi
#2}}
\providecommand{\BIBdecl}{\relax}
\BIBdecl

\bibitem{wang2020language}
C.~Wang, X.~Liu, and D.~Song, ``Language models are open knowledge graphs,''
  \emph{arXiv preprint arXiv:2010.11967}, 2020.

\bibitem{gao2023retrieval}
Y.~Gao, Y.~Xiong, X.~Gao, K.~Jia, J.~Pan, Y.~Bi, Y.~Dai, J.~Sun, and H.~Wang,
  ``Retrieval-augmented generation for large language models: A survey,''
  \emph{arXiv preprint arXiv:2312.10997}, 2023.

\bibitem{asai2023self}
A.~Asai, Z.~Wu, Y.~Wang, A.~Sil, and H.~Hajishirzi, ``Self-rag: Learning to
  retrieve, generate, and critique through self-reflection,'' \emph{arXiv
  preprint arXiv:2310.11511}, 2023.

\bibitem{gao2022rarr}
L.~Gao, Z.~Dai, P.~Pasupat, A.~Chen, A.~T. Chaganty, Y.~Fan, V.~Y. Zhao,
  N.~Lao, H.~Lee, D.-C. Juan \emph{et~al.}, ``Rarr: Researching and revising
  what language models say, using language models,'' \emph{arXiv preprint
  arXiv:2210.08726}, 2022.

\bibitem{kuzi2020leveraging}
S.~Kuzi, M.~Zhang, C.~Li, M.~Bendersky, and M.~Najork, ``Leveraging semantic
  and lexical matching to improve the recall of document retrieval systems: A
  hybrid approach,'' \emph{arXiv preprint arXiv:2010.01195}, 2020.

\bibitem{zhang2016neural}
Y.~Zhang, M.~M. Rahman, A.~Braylan, B.~Dang, H.-L. Chang, H.~Kim, Q.~McNamara,
  A.~Angert, E.~Banner, V.~Khetan \emph{et~al.}, ``Neural information
  retrieval: A literature review,'' \emph{arXiv preprint arXiv:1611.06792},
  2016.

\bibitem{khattab2020colbert}
O.~Khattab and M.~Zaharia, ``Colbert: Efficient and effective passage search
  via contextualized late interaction over bert,'' in \emph{Proceedings of the
  43rd International ACM SIGIR conference on research and development in
  Information Retrieval}, 2020, pp. 39--48.

\bibitem{liu2307lost}
N.~F. Liu, K.~Lin, J.~Hewitt, A.~Paranjape, M.~Bevilacqua, F.~Petroni, and
  P.~Liang, ``Lost in the middle: How language models use long contexts,''
  \emph{Transactions of the Association for Computational Linguistics},
  vol.~12, pp. 157--173, 2024.

\bibitem{kaddour2023challenges}
J.~Kaddour, J.~Harris, M.~Mozes, H.~Bradley, R.~Raileanu, and R.~McHardy,
  ``Challenges and applications of large language models,'' \emph{arXiv
  preprint arXiv:2307.10169}, 2023.

\bibitem{es2023ragas}
S.~Es, J.~James, L.~Espinosa-Anke, and S.~Schockaert, ``Ragas: Automated
  evaluation of retrieval augmented generation,'' \emph{arXiv preprint
  arXiv:2309.15217}, 2023.

\bibitem{chen2024benchmarking}
J.~Chen, H.~Lin, X.~Han, and L.~Sun, ``Benchmarking large language models in
  retrieval-augmented generation,'' in \emph{Proceedings of the AAAI Conference
  on Artificial Intelligence}, vol.~38, no.~16, 2024, pp. 17\,754--17\,762.

\bibitem{goswami2023switchprompt}
K.~Goswami, L.~Lange, J.~Araki, and H.~Adel, ``Switchprompt: Learning
  domain-specific gated soft prompts for classification in low-resource
  domains. corr, abs/2302.06868,'' 2023.

\bibitem{guu2020retrieval}
K.~Guu, K.~Lee, Z.~Tung, P.~Pasupat, and M.~Chang, ``Retrieval augmented
  language model pre-training,'' in \emph{International conference on machine
  learning}.\hskip 1em plus 0.5em minus 0.4em\relax PMLR, 2020, pp. 3929--3938.

\bibitem{lewis2020retrieval}
P.~Lewis, E.~Perez, A.~Piktus, F.~Petroni, V.~Karpukhin, N.~Goyal,
  H.~K{\"u}ttler, M.~Lewis, W.-t. Yih, T.~Rockt{\"a}schel \emph{et~al.},
  ``Retrieval-augmented generation for knowledge-intensive nlp tasks,''
  \emph{Advances in Neural Information Processing Systems}, vol.~33, pp.
  9459--9474, 2020.

\bibitem{luan2021sparse}
Y.~Luan, J.~Eisenstein, K.~Toutanova, and M.~Collins, ``Sparse, dense, and
  attentional representations for text retrieval,'' \emph{Transactions of the
  Association for Computational Linguistics}, vol.~9, pp. 329--345, 2021.

\bibitem{wang2021retrieval}
Z.~Wang, P.~Ng, R.~Nallapati, and B.~Xiang, ``Retrieval, re-ranking and
  multi-task learning for knowledge-base question answering,'' in
  \emph{Proceedings of the 16th Conference of the European Chapter of the
  Association for Computational Linguistics: Main Volume}, 2021, pp. 347--357.

\bibitem{juvekar2024introducing}
K.~Juvekar and A.~Purwar, ``Introducing a new hyper-parameter for rag: Context
  window utilization,'' \emph{arXiv preprint arXiv:2407.19794}, 2024.

\bibitem{finardi2024chronicles}
P.~Finardi, L.~Avila, R.~Castaldoni, P.~Gengo, C.~Larcher, M.~Piau, P.~Costa,
  and V.~Carid{\'a}, ``The chronicles of rag: The retriever, the chunk and the
  generator,'' \emph{arXiv preprint arXiv:2401.07883}, 2024.

\bibitem{llama-index}
\emph{LlamaIndex: \url{https://docs.llamaindex.ai/en/stable/}}.

\bibitem{oai-evals}
\emph{OpenAI Evals: \url{https://github.com/openai/evals}}.

\bibitem{zhu2024rageval}
K.~Zhu, Y.~Luo, D.~Xu, R.~Wang, S.~Yu, S.~Wang, Y.~Yan, Z.~Liu, X.~Han, Z.~Liu
  \emph{et~al.}, ``Rageval: Scenario specific rag evaluation dataset generation
  framework,'' \emph{arXiv preprint arXiv:2408.01262}, 2024.

\bibitem{yao2022react}
S.~Yao, J.~Zhao, D.~Yu, N.~Du, I.~Shafran, K.~Narasimhan, and Y.~Cao, ``React:
  Synergizing reasoning and acting in language models,'' \emph{arXiv preprint
  arXiv:2210.03629}, 2022.

\bibitem{marvin2023prompt}
G.~Marvin, N.~Hellen, D.~Jjingo, and J.~Nakatumba-Nabende, ``Prompt engineering
  in large language models,'' in \emph{International conference on data
  intelligence and cognitive informatics}.\hskip 1em plus 0.5em minus
  0.4em\relax Springer, 2023, pp. 387--402.

\bibitem{wei2022chain}
J.~Wei, X.~Wang, D.~Schuurmans, M.~Bosma, F.~Xia, E.~Chi, Q.~V. Le, D.~Zhou
  \emph{et~al.}, ``Chain-of-thought prompting elicits reasoning in large
  language models,'' \emph{Advances in neural information processing systems},
  vol.~35, pp. 24\,824--24\,837, 2022.

\bibitem{yao2023tree}
S.~Yao, D.~Yu, J.~Zhao, I.~Shafran, T.~L. Griffiths, Y.~Cao, and K.~Narasimhan,
  ``Tree of thoughts: Deliberate problem solving with large language models,
  2023,'' \emph{URL https://arxiv. org/pdf/2305.10601. pdf}, 2023.

\bibitem{zhang2019bertscore}
T.~Zhang, V.~Kishore, F.~Wu, K.~Q. Weinberger, and Y.~Artzi, ``Bertscore:
  Evaluating text generation with bert,'' \emph{arXiv preprint
  arXiv:1904.09675}, 2019.

\bibitem{li2022faithfulness}
W.~Li, W.~Wu, M.~Chen, J.~Liu, X.~Xiao, and H.~Wu, ``Faithfulness in natural
  language generation: A systematic survey of analysis, evaluation and
  optimization methods,'' \emph{arXiv preprint arXiv:2203.05227}, 2022.

\bibitem{lin2023llm}
Y.-T. Lin and Y.-N. Chen, ``Llm-eval: Unified multi-dimensional automatic
  evaluation for open-domain conversations with large language models,''
  \emph{arXiv preprint arXiv:2305.13711}, 2023.

\bibitem{zhong2024mix}
Z.~Zhong, H.~Liu, X.~Cui, X.~Zhang, and Z.~Qin, ``Mix-of-granularity: Optimize
  the chunking granularity for retrieval-augmented generation,'' \emph{arXiv
  preprint arXiv:2406.00456}, 2024.

\bibitem{zhao2024dense}
W.~X. Zhao, J.~Liu, R.~Ren, and J.-R. Wen, ``Dense text retrieval based on
  pretrained language models: A survey,'' \emph{ACM Transactions on Information
  Systems}, vol.~42, no.~4, pp. 1--60, 2024.

\bibitem{reimers2019sentence}
N.~Reimers, ``Sentence-bert: Sentence embeddings using siamese bert-networks,''
  \emph{arXiv preprint arXiv:1908.10084}, 2019.

\bibitem{gao2021simcse}
T.~Gao, X.~Yao, and D.~Chen, ``Simcse: Simple contrastive learning of sentence
  embeddings,'' \emph{arXiv preprint arXiv:2104.08821}, 2021.

\bibitem{tay2022transformer}
Y.~Tay, V.~Tran, M.~Dehghani, J.~Ni, D.~Bahri, H.~Mehta, Z.~Qin, K.~Hui,
  Z.~Zhao, J.~Gupta \emph{et~al.}, ``Transformer memory as a differentiable
  search index,'' \emph{Advances in Neural Information Processing Systems},
  vol.~35, pp. 21\,831--21\,843, 2022.

\bibitem{bge-base-en-v1.5}
\emph{\url{https://huggingface.co/BAAI/bge-base-en-v1.5}}.

\bibitem{ms-marco-MiniLM-L-2-v2}
\emph{\url{https://huggingface.co/cross-encoder/ms-marco-MiniLM-L-2-v2}}.

\bibitem{dubey2024llama}
A.~Dubey, A.~Jauhri, A.~Pandey, A.~Kadian, A.~Al-Dahle, A.~Letman, A.~Mathur,
  A.~Schelten, A.~Yang, A.~Fan \emph{et~al.}, ``The llama 3 herd of models,''
  \emph{arXiv preprint arXiv:2407.21783}, 2024.

\bibitem{qwen2.5}
\BIBentryALTinterwordspacing
Q.~Team, ``Qwen2.5: A party of foundation models,'' September 2024. [Online].
  Available: \url{https://qwenlm.github.io/blog/qwen2.5/}
\BIBentrySTDinterwordspacing

\end{thebibliography}

\end{document}